\documentclass[letterpaper, 10 pt, conference]{ieeeconf}  

\IEEEoverridecommandlockouts                              

\usepackage{graphicx} 
\usepackage{caption}
\usepackage{subcaption}
\usepackage{pgf}
\usepackage{tikz}
\usepackage{mathtools} 
\usepackage{amssymb}  
\usepackage{hyperref}
\usepackage{siunitx}
\usepackage{comment}
\usepackage{subcaption}
\usepackage{eso-pic}

\let\labelindent\relax
\usepackage[inline]{enumitem}

\usetikzlibrary{arrows,automata,shapes,fit,positioning}

\newlength\innermarg
\newlength\outerlinewidth
\renewcommand{\vec}[1]{\boldsymbol{#1}} 

\newcommand{\acronym}[1]{\textsc{\MakeLowercase{#1}}} 

\newcommand{\norm}[1]{\lVert#1\rVert_2}

\newcommand\circled[1]{\tikzset{external/export next=false}\tikz[baseline=(char.base)]{
            \node[shape=circle,draw,fill=black,text=white,inner sep=0pt] (char) {\textbf{#1}};}}

\makeatletter
\newcommand\resetstackedplots{
\makeatletter
\pgfplots@stacked@isfirstplottrue
\makeatother
\addplot [forget plot,draw=none] coordinates{(1,0) (2,0) (3,0)};
}
\makeatother

\usetikzlibrary{arrows,arrows,arrows.meta,automata,calc,decorations.pathreplacing,external,fit,pgfplots.groupplots,positioning,shapes}

\newcommand\copyrighttext{%
  \footnotesize \textcopyright 2025 IEEE. Personal use of this material is permitted.
  Permission from IEEE must be obtained for all other uses, in any current or future
  media, including reprinting/republishing this material for advertising or promotional
  purposes, creating new collective works, for resale or redistribution to servers or
  lists, or reuse of any copyrighted component of this work in other works.
}

\newcommand\copyrightnotice[1][black]{%
\begin{tikzpicture}[remember picture,overlay]
\node[anchor=south,yshift=10pt,draw=#1] at (current page.south) {\parbox{\dimexpr\textwidth-\fboxsep-\fboxrule\relax}{\copyrighttext}};
\end{tikzpicture}%
}

\newcommand\overlaycopyrightnotice[1][black]{%
\AddToShipoutPicture*{\copyrightnotice[#1]}%
}

\title{\LARGE \bf
Sampling-Based Grasp and Collision Prediction for Assisted Teleoperation
}

\author{Simon Manschitz$^{1}$ and Berk Gueler$^{1,2}$ and Wei Ma$^{1}$ and Dirk Ruiken$^{1}$
\thanks{$^{1}$All authors are with Honda Research Institute Europe GmbH, Carl-Legien-Straße 30, 63073 Offenbach/Main, Germany
        {\tt\small simon.manschitz@honda-ri.de}}%
\thanks{$^{2}$Berk Gueler is with the Institute for Intelligent Autonomous Systems, Technische Universit\"at Darmstadt, 64289 Darmstadt, Germany}}

\begin{document}

\maketitle
\overlaycopyrightnotice
\thispagestyle{empty}
\pagestyle{empty}

\begin{abstract}
Shared autonomy allows for combining the global planning capabilities of a human operator with the strengths of a robot such as repeatability and accurate control. In a real-time teleoperation setting, one possibility for shared autonomy is to let the human operator decide for the rough movement and to let the robot do fine adjustments, e.g., when the view of the operator is occluded.
We present a learning-based concept for shared autonomy that aims at supporting the human operator in a real-time teleoperation setting. At every step, our system tracks the target pose set by the human operator as accurately as possible while at the same time satisfying a set of constraints which influence the robot's behavior. An important characteristic is that the constraints can be dynamically activated and deactivated which allows the system to provide task-specific assistance.
Since the system must generate robot commands in real-time, solving an optimization problem in every iteration is not feasible. Instead, we sample potential target configurations and use Neural Networks for predicting the constraint costs for each configuration. 
By evaluating each configuration in parallel, our system is able to select the target configuration which satisfies the constraints and has the minimum distance to the operator's target pose with minimal delay.
We evaluate the framework with a pick and place task on a bi-manual setup with two Franka Emika Panda robot arms with Robotiq grippers.
\end{abstract}

\section{Introduction}
Teleoperation allows a human operator to remotely control a robot.
Such a control scheme is especially helpful when the human cannot be present at the remote location, for instance due to a hazardous environment.
However, it is often difficult for a human operator to control a robot remotely. 
A robot manipulator has many degrees of freedom~(\acronym{DOF}) and controlling them individually is difficult and time-consuming. 
Therefore, various methods have been proposed~(e.g., \cite{ThompsonICORR2022,BissonRACISS2015, WangICRA2021, QuereICRA2020}) to automatically remap the human input to a robot command, making the overall teleoperation more intuitive for the operator. 
One example of remapping is to let the operator only control the kinematic pose of the end-effector, for instance with a joystick or gamepad-like controller. In that case, the target of the system becomes to track a virtual target set by the human operator and the \acronym{DOFs} of the robot can be determined by a standard Inverse Kinematics solver. Even with an intuitive mapping of the human input to robot control commands, teleoperation can be difficult for the operator due to perceptual issues or the lack of haptic feedback. However, incorporating haptic feedback devices into teleoperation, while beneficial in some contexts, often introduces complexities and costs. On the other hand, haptic feedback can lead to sensory conflicts, particularly when combined with visual input ~\cite{SelvaggioRAS2021}. At the same time, completely autonomous execution of manipulation tasks is usually not possible due to their highly dynamic nature.
Shared autonomy is a research field which aims at alleviating some of the aforementioned issues~\cite{SelvaggioRAS2021, MowerROMAN2019}. In shared autonomy, the human operator is not controlling the robot all the time. Instead, the system can take over control during certain phases of a task. For instance, it may take over control in order to prevent the robot from being harmed, or it may take over control in a phase of the task that can easily be solved because of its simplicity.

\begin{figure}[t]
  \includegraphics[width=\linewidth]{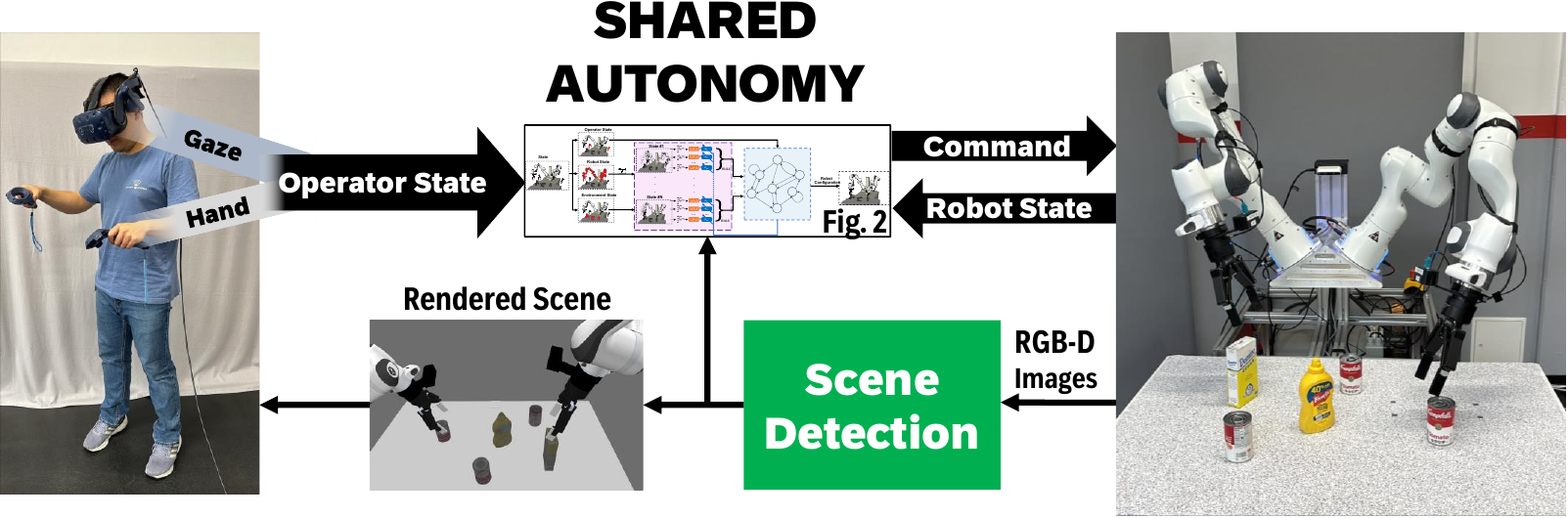}
  \centering
  \caption{The system extracts a scene from raw \acronym{RGB-D} images and displays it to the operator.
  It processes the scene together with other input~(operator hand state, gaze, and the robot state) and generates a command for the robot. In contrast to an unassisted teleoperation setup, the system estimates the operator's intention and may correct the operator's input commands in certain situations. A detailed overview of the shared autonomy module is provided in \hyperref[fig:framework]{Figure~\ref*{fig:framework}}.}
  \label{fig:overview}
\end{figure}

In this paper, we present a shared autonomy framework designed for real-time assisted teleoperation. Therefore, the terms shared autonomy and assisted teleoperation will be used interchangeably. We define teleoperation as tracking a target pose set continuously by the human operator. As depicted in \hyperref[fig:overview]{Figure~\ref*{fig:overview}}, the operator wears a head mounted display and controls the robot's end-effectors with a pair of controllers. The operator sees a rendered virtual reality~(\acronym{VR}) scene which includes the detected objects, the robot and the target pose.
The aim of the shared autonomy framework is to enable the operator to perform a manipulation task with a remotely controlled robot while preserving their sense of agency — that is, their feeling of being in control of the system. The core idea is to ensure that the robot’s end-effectors follow the target poses set by the operator as closely as possible while simultaneously satisfying a set of task-dependent constraints that may change over time. These constraints enhance the safety and robustness of teleoperation, representing the assistance aspect of the framework.

Examples for constraints are joint limit avoidance, obstacle collisions, self-collisions or pre-grasp pose validity. As an example, consider a simple pick task. The robot starts in an initial configuration far away from an object. The human operator moves the target pose close to the object s(he) wants to grasp and the robot approaches the object. During this phase, the robot should follow the target pose as closely as possible while avoiding joint limits and collisions with itself and the environment (e.g., static objects such as a table). When being close enough to the object, the "pre-grasp constraint" is activated, which means from now on the robot end-effector should reach a pose where it becomes possible to grasp the object. It is important to note the task is not to find an optimal pre-grasp pose, but to find one which is as closely as possible (according to a distance metric) to the target pose set by the human operator. Therefore, the human operator retains the sense of agency and is still able to do fine-adjustments of the pre-grasp pose. When the operator is satisfied with the pre-grasp pose of the robot, s(he) can close the gripper, e.g., by pressing a button the controller and the object is grasped by the robot.

The contribution of this paper is two-fold. First, we present a concept for locally approximating the active constraint costs around the current robot configuration. Second, we show how that information can be utilized for deciding in which direction to move the robot such that the active constraints are satisfied and the difference to the target pose set by the human operator is minimized, if possible. We evaluate our approach on a bi-manual robot setup with two Franka Emika Panda arms with Robotiq grippers, showing that it can be utilized for providing assistance when teleoperating a real robot in real-time.

\section{Related Work}
In shared autonomy, one challenge is to find a good balance between adequate assistance and retaining the operator's sense of agency.
Achieving such a balance is crucial for shared autonomy systems, and various methods have been discussed in the literature. These methods often involve leveraging the human operator's intention, extracting environmental information, or take into account specific task requirements. 
We categorize the related work into human intention-based methods, which primarily revolve around human intent estimation as their core component and environment/task-based methods.

\textbf{Human Intention-Based Methods}:
In human-in-the-loop systems, understanding the human's intention as well as knowing the current manipulation target are required in order to support human operators. One possibility is to let the operator manually define the current target action through a user interface~\cite{TheofanidisPETRA2017, Park1991}. Developing comprehensive user interfaces may not always be possible or desirable. Therefore, human intention can also be inferred automatically, e.g., through gaze inputs in a \acronym{VR} setup~\cite{FuchsFrNeu2021, WangICRA2021,WangICACI2020}, gestures~\cite{vanc2023communicating}, joystick inputs~\cite{Jain2018}, speech~\cite{WangICACI2020} or based on the operator's hand trajectory~\cite{GamageUIST2021}.

In~\cite{WangICRA2021}, an approach is presented that utilizes the operator's intention to autocorrect the commands sent to the robot. Similar approaches are presented in~\cite{GonzalesJOMS2021,Manschitz2022,ZeinICRA2021}. 
Instead of focusing solely on intention prediction, the authors of~\cite{Dragan2013} propose a human-centric approach. By training human operators to provide input signals that are easier for the system to interpret, the authors enable operators to communicate their intent more transparently. In~\cite{jeon2020shared}, the authors reduce the complexity of remapping human input by using latent actions. They propose a way to embed human goals and propose precise assistive dexterous manipulation utilizing shared autonomy and learned latent actions.

\textbf{Environment/Task-Based Methods}: 
Within the realm of shared autonomy, a distinct area of research centers on approaches that emphasize the influence of the environment and the goals associated with specific tasks. These methods prioritize the adaptation of robot actions based on contextual factors within the environment and the precise objectives of the task in question~\cite{BrooksJournal2024, ArpinoJournal2024}. In ~\cite{Reddy2018}, a deep reinforcement learning policy was trained using task-specific rewards based on observations of the environment and user input. When the user inputs are optimal, the trained policy closely follows the user's commands. However, when they are not optimal, the policy deviates from the user's input in order to maximize the reward extracted from the environment. In addition, in \cite{ewerton2019}, the authors proposed a reinforcement learning algorithm that enables a  teleoperated robot to adapt to changes in the environment.
In~\cite{BirkenkampfDLR2014}, the authors propose an intuitive interface for shared autonomy designed for use in predefined environments, enabling users to issue high-level commands. 

In this paper, we employ the methodology outlined in~\cite{FuchsFrNeu2021} to anticipate the operator's intentions. Roughly speaking, when the operator moves the target pose towards an object and/or the operator's gaze is directed toward that object, it will be chosen as target object. Compared to many of the aforementioned approaches, we believe our constraint-based formulation offers the advantage of being easily adaptable to new tasks, thanks to its modular design. For example, supporting a different manipulation task may require introducing new task-specific constraints, but existing constraints—such as those for collision avoidance or grasping—can often be reused, making the system more flexible and efficient.

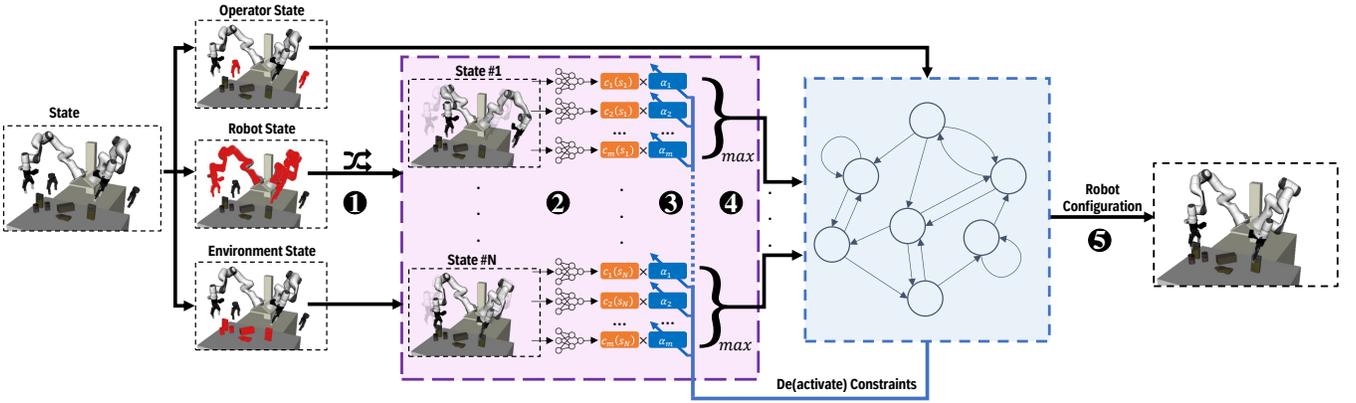
\begin{figure*}[t]
  \begin{tikzpicture}[inner sep=0, outer sep=0]
    \node at (0,0) {\includegraphics[width=\linewidth]{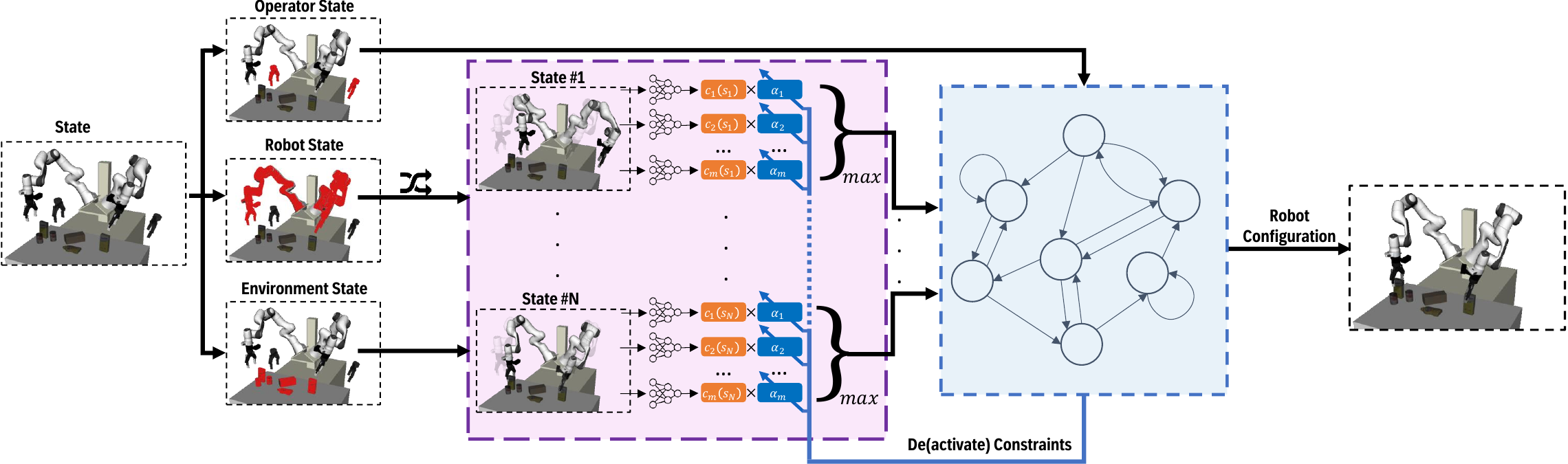}}; 
    \node[shape=circle,draw,fill=black,text=white,inner sep=0pt] at (-4.2, +0.0) (char) {\textbf{1}};
    \node[shape=circle,draw,fill=black,text=white,inner sep=0pt] at (-1.5, +0.0) (char) {\textbf{2}};
    \node[shape=circle,draw,fill=black,text=white,inner sep=0pt] at (+0.0, +0.0) (char) {\textbf{3}};
    \node[shape=circle,draw,fill=black,text=white,inner sep=0pt] at (+0.8, +0.0) (char) {\textbf{4}};
    \node[shape=circle,draw,fill=black,text=white,inner sep=0pt] at (+5.7, -0.5) (char) {\textbf{5}};
  \end{tikzpicture}  
  \centering
  \caption{Overview of our framework. \protect\circled{1} We randomly sample robot states close to the current state and \protect\circled{2} predict the constraint costs~$c_i$ for each sampled state. The predicted costs are then \protect\circled{3} multiplied by factors $\alpha_i$ that are dynamically controlled by a state machine. Next, the costs are \protect\circled{4} aggregated and fed into said state machine where a target configuration is selected from the samples and \protect\circled{5} sent as command to the robot. More details can be found in \hyperref[sec:framework]{Section~\ref*{sec:framework}}.}
  \label{fig:framework}
\end{figure*}

\section{Assisted Teleoperation Framework}\label{sec:framework}
An overview of our framework is depicted in \hyperref[fig:framework]{Figure~\ref*{fig:framework}}. We assume the full state $\vec{s} \in \mathbb{R}^{D}$ is comprised of the robot's joint configuration  $\vec{q} \in \mathbb{R}^{N_J}$ where $N_J$ is the number of joints, the robot's end-effector pose $\vec{s}_\mathrm{eff} \in \mathbb{R}^6$, the object poses $\vec{s}_{o} \in \mathbb{R}^{6 \times N_O}$, where $N_O$ is the number of objects. The state $\vec{s}_{h} \in \mathbb{R}^7$ of the human operator is comprised of the $6D$ target pose and the desired gripper state which is a continuous value $[0, 1]$. Since our setup is bi-manual, the human operator state and the end-effector pose are defined for the left and right hand, respectively. The aim of our shared autonomy framework is to minimize the distance between the robot's end-effector $\vec{s}_\mathrm{eff}(\vec{q})$ and the operator's target pose $\vec{s}_\mathrm{h}$ under consideration of the active constraints. More formally, we can define the optimization problem as
\begin{align}\label{eq:optimization}
    &\underset{\vec{q}}{\text{min}}&&\norm{f(\vec{s}^{\mathrm{left}}_{\mathrm{eff}}(\vec{q}), \vec{s}^{\mathrm{left}}_\mathrm{h})}^2 + \norm{f(\vec{s}^{\mathrm{right}}_{\mathrm{eff}}(\vec{q}), \vec{s}^{\mathrm{right}}_\mathrm{h})}^2 \notag \\
    &\text{subject to} && \alpha_1 c_1(\vec{s}) \leq 0, \dots, \alpha_\mathrm{N_c} c_\mathrm{N_c}(\vec{s}) \leq 0.  
\end{align}
As distance function $f$ we use a weighted sum of the Euclidean distance of the positions and the rotation angle between the orientations. The factors $\alpha_i$ can be utilized for activating ($\alpha_i=1$) or deactivating ($\alpha_i=0$) a constraint $c_i$ during execution.

Since we aim at real-time teleoperation, the joint command~$\vec{q}$ should be updated and sent to the controller at a rate of roughly $\SI{25}{\hertz}$ without a noticeable delay. Therefore, solving the optimization problem~\ref{eq:optimization} is infeasible. Instead, we propose to locally approximate the optimization problem and let the robot end-effector move in a direction which minimizes the constraint costs (if the constraints are violated) and the distance to the target pose at the same time. For speeding up the computation, we first sample a set of robot configurations close to the current robot state $\Delta\vec{q}_1, \Delta\vec{q}_2, \dots, \Delta\vec{q}_\mathrm{N_s}$ where $N_s$ is the number of samples (1024 in this paper). Next, the constraint costs for each active constraint are predicted for each sampled configuration. Finally, one of the sampled configurations is selected as next target pose for the robot based on the predicted constraint costs and the distance to the target pose of that configuration. In the following sections, we provide details about the individual parts of the framework and how it can be utilized for robot teleoperation under real-time constraints. An overview of the framework is depicted in~\hyperref[fig:framework]{Figure~\ref*{fig:framework}}.

\subsection{Constraint Costs Prediction}\label{sec:constraints}
As mentioned before, each manipulation tasks has a set of specific constraints a robot must satisfy in order to successfully perform the task. 
For instance, when moving a grasped object in free space a certain pose might be required, e.g. a cup of water must be held such that the water is not spilled. Our interpretation of shared autonomy in teleoperation is that the system should ensure the constraints are satisfied, while the operator can freely move the robot in all directions that do not lead to a constraint violation.
One prerequisite for using constraints in our real-time teleoperation setting is that we must be able to compute the constraint costs efficiently. While some constraints can be manually defined~(e.g., pose constraint for an object), others must be approximated through a model since manually defining them would be too time-consuming or difficult. For instance, valid (pre-)grasp poses are difficult or time-consuming to model, but it is possible to learn them from examples~\cite{NewburyTRO2023, PlattARC2023}. In this paper, we assume that we can define $N_C$ different constraints according to our task. The constraint cost of constraint $i$ is referred to as $c_i(\vec{s})$. Here, $\vec{s}$ corresponds to the current state of the environment~(e.g., robot configuration and object poses). A constraint $i$ is satisfied for $c_i(\vec{s}) \leq 0$ and violated otherwise. Since we only approximate the constraint costs, we can add a small safety margin to minimize the likelihood of false negatives~$c_i(\vec{s}) \approx \Tilde{c}_i(\vec{s}) \leq \delta$. The safety margin enables fine-tuning of individual constraints. For example, increasing the safety margin for collision constraints can help minimize false negative collision predictions, thereby enhancing overall robot safety.

In this paper, each constraint is approximated with a standard Feedforward Neural Network with fully connected layers and {\it LeakyReLU} activation functions, but in general any type of model could be used. We also tested other activation functions, such as {\it ReLU} and {\it Softplus}, but did not observe any significant performance differences. One argument for predicting the constraint costs instead of computing them is that many constraints are naturally either satisfied or violated, resulting in a binary constraint cost value (e.g., $\pm1$). When approximating such constraints, the predicted constraints costs become continuous and give the system some notion of how far the configuration is away from the violation border, yielding in richer information. Another advantage is that Neural Networks can easily predict the constraint costs for multiple robot configurations in parallel, allowing to speed up the computation in our case.

\subsection{Sampling-based Target Configuration Selection}\label{sec:direction}
During runtime, the predicted constraint costs for each sampled configuration are evaluated. For each sample, the individual constraint costs are aggregated by taking the maximum value of all activate constraints. The consequence is that as soon as at least one constraint is violated for a given configuration, the aggregated value is also positive. In this paper, we measure the target distance~$\norm{f(\vec{s}_{\mathrm{eff}}(\vec{q}), \vec{s}_\mathrm{h})}^2$~(see~\ref{eq:optimization}) for each configuration and select the configuration which has the smallest target distance and satisfies all activate constraints as next target configuration for the robot. If no feasible configuration is found, the target configuration is not updated and the hence the robot will stop moving. Since the resulting robot motion might be a bit jerky, we generate a sequence of target configurations~($3$ in this paper) and filter the resulting joint angle trajectory with a Hanning window. Before sending the command to the robot, we compute the constraint costs for the filtered target configuration to verify that it does not violate the constraint. Usually, that is not the case since the target configurations are generated at a relatively high rate and the delta for the joint randomization is small.

\subsection{Dynamic Constraint Activation}
During teleoperation, we must decide which constraints to activate in which situation. In this paper, we focus on picking up objects. Therefore, we defined five different constraints. {\it Static collisions}: satisfied if the robot arm is not in a collision with static objects in the environment (e.g., a table). {\it Self collisions}: satisfied if the robot arm is not in a collision with itself. {\it Dynamic collisions}: satisfied if the robot arm is not in a collision with dynamic objects in the environment (e.g., objects which are graspable). {\it Mutual collisions}: satisfied if the robot arms are not colliding with each other. {\it Pre-grasp}: satisfied if the current robot configuration allows the robot gripper to grasp an object. All constraints except the {\it mutual collisions} predict the constraint for a single arm. Since our setup is comprised of two robot arms, we have nine constraints in total.

The constraints are dynamically activated and deactivated with a simple state machine. The robot starts in an initial state called {\it teleoperation} where all collision constraints are activated while the {\it pre-grasp} constraint is deactivated. Hence, the desired behavior in this state is that the robot should follow the target pose set by the human operator without colliding with objects. Joint limits are always met by the controller by clipping the joint commands. When the robot's end-effector approaches an object and the (predicted) {\it pre-grasp} constraint is satisfied for at least one sampled robot configuration, the constraint is activated and the system switches to a behavior we call {\it Align With Object}. In this state, all constraints are activated which means the robot will automatically move the gripper to a feasible grasp pose. The desired behavior in this state is that the operator should be able to move the end-effector around the object. At each point in time, the current robot configuration should correspond to a valid pre-grasp pose which means the state allows the operator to adjust the pre-grasp pose. When pressing a button on the controller, the gripper is closed and the object should be grasped by the robot. In this paper, after grasping an object, no assistance is provided until the object has been placed down. We consider it future work to also provide such a functionality.

\subsection{Object Detection}
For object detection and object pose estimation, the scene is recorded with two static, calibrated Intel RealSense \acronym{RGB-D} cameras. Images are segmented with one instance of \acronym{Mask R-CNN}~\cite{HeICCV2017} per camera. Point clouds are generated for each segment and fused across cameras. Feature are extracted~\cite{RusuICRA2009} followed by global registration~\cite{ZhouECCV2016} and fine registration~\cite{ChenIVC1992}. Finally, pose confidence estimation (similar to visible surface discrepancy in~\cite{HodavnECCV2016}) is used to filter the results. Whenever the robot end-effector approaches an object we freeze the pose of that object in order to prevent the object from disappearing from the scene due to occlusion.

\section{Experiments}\label{sec:experiments}
\begin{figure}[t]
  \centering
  \begin{tikzpicture}[node distance=0.1cm, inner sep=0pt, nodes={draw, thin}]
    \node[]                       (pt1) {\includegraphics[width=.475\linewidth]{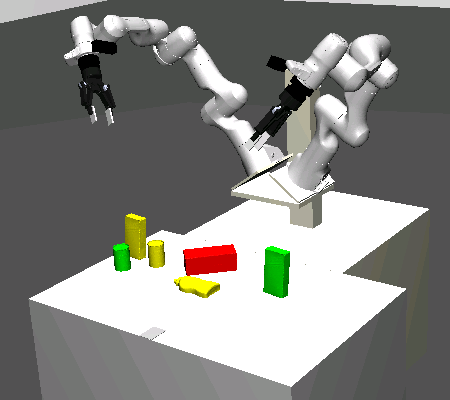}};
    \node[anchor=west, right=of pt1.east]   (pt2) {\includegraphics[width=.475\linewidth]{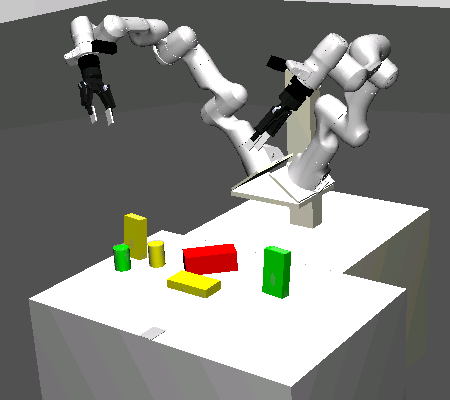}};
  \end{tikzpicture}    
  \caption{Snapshots of environment used for generating the training data. The pictures show objects from the \acronym{YCB}-video dataset (left) and their box- or cylinder approximations used for predicting collisions and feasible grasp poses (right).}
  \label{fig:setup}
\end{figure}

We evaluate our method on a bi-manual setup with two Franka Emika Panda robot arms with Robotiq grippers, as depicted in \hyperref[fig:overview]{Figure~\ref*{fig:overview}}. We evaluate our method on a pick and place task on the real robot and compare it against a baseline that does not provide any assistance.

\textbf{Dataset Generation}:
The datasets for training the constraint cost predictors were generated in simulation, as depicted in \hyperref[fig:setup]{Figure~\ref*{fig:setup}}. In this paper, we concentrate on picking box- and cylinder-shaped objects. Therefore, we randomly placed boxes and cylinders on the table. We also randomized the sizes of the objects (e.g., the cylinder radius was varied between $\SI{1}{cm}$ and $\SI{5}{cm}$).
Next, we randomized the robot joint configuration and computed the constraint costs. We generated an individual dataset for each constraint listed in \hyperref[tab:constraints]{Table~\ref*{tab:constraints}}. In order to get a balanced dataset, we sampled configurations until $50\%$ of the samples satisfied the constraints. Each training dataset contained 524288 samples, while the test datasets contained 16384 samples.

\begin{table}[h!]
  \centering
  \begin{tabular}{| c | c@{\hspace{2pt}}c@{\hspace{2pt}}c | c c|} 
   \hline  
   & \multicolumn{3}{c|}{Model Details} & \multicolumn{2}{c|}{Accuracy [\%]} \\ 
   Constraint & Features & Layers & Params & Train & Test \\ 
   \hline
   Static Collisions l+r & 8 & 7 & 331521 & 98.8 & 97.7 \\ 
   Mutual Collisions & 16 & 7 & 333569 & 98.1 & 97.2 \\ 
   Self Collisions l+r & 8 & 7 & 331521 & 99.3 & 99.0 \\ 
   Dynamic Collisions l+r & 13 & 7 & 332801 & 96.5 & 93.6 \\
   Valid pre-grasp l+r & 7 & 5 & 331265 & 97.5 & 96.6 \\
   \hline
  \end{tabular}
  \caption{Overview over different constraint cost predictors with the characteristics of the learned models as well as training and testing accuracy. All predictors except for the mutual collision predictor were trained for left and right arm, separately.}
  \label{tab:constraints}
\end{table}

\begin{figure*}
  \centering
  \begin{tikzpicture}[node distance=0.1cm, inner sep=0pt, nodes={draw, thin}]
    \node[]                                (pt1) {\includegraphics[width=.19\textwidth]{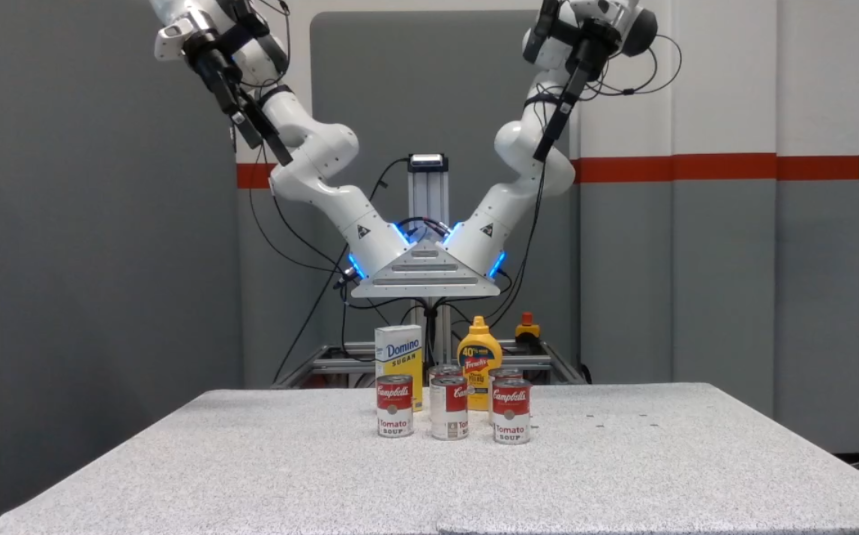}};
    \node[anchor=west, right=of pt1.east]  (pt2) {\includegraphics[width=.19\textwidth]{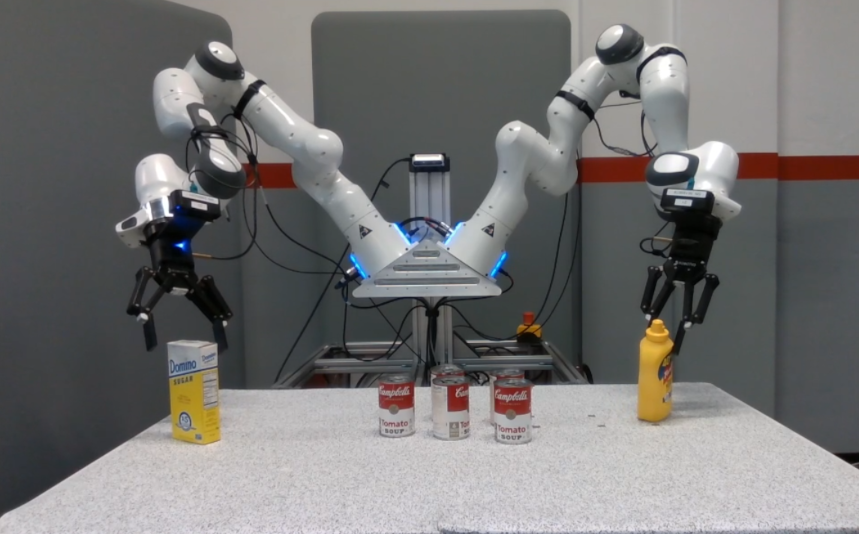}};
    \node[anchor=west, right=of pt2.east]  (pt3) {\includegraphics[width=.19\textwidth]{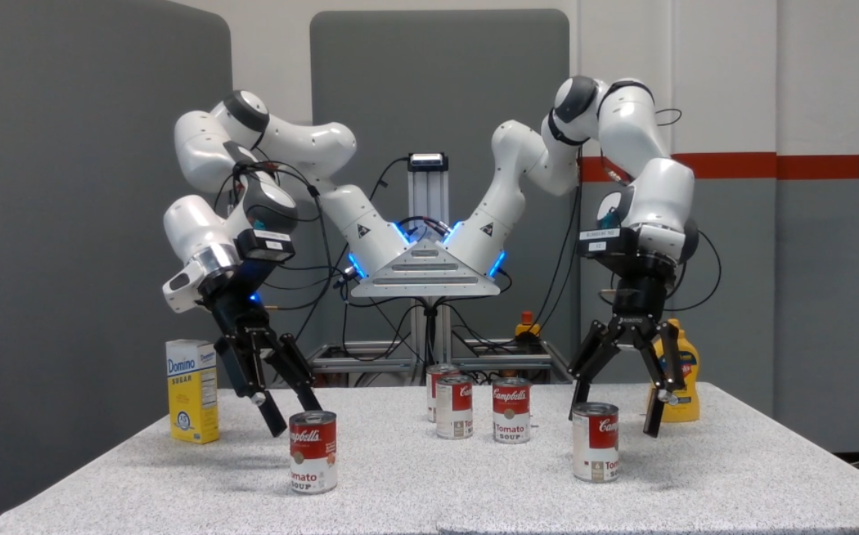}};
    \node[anchor=west, right=of pt3.east]  (pt4) {\includegraphics[width=.19\textwidth]{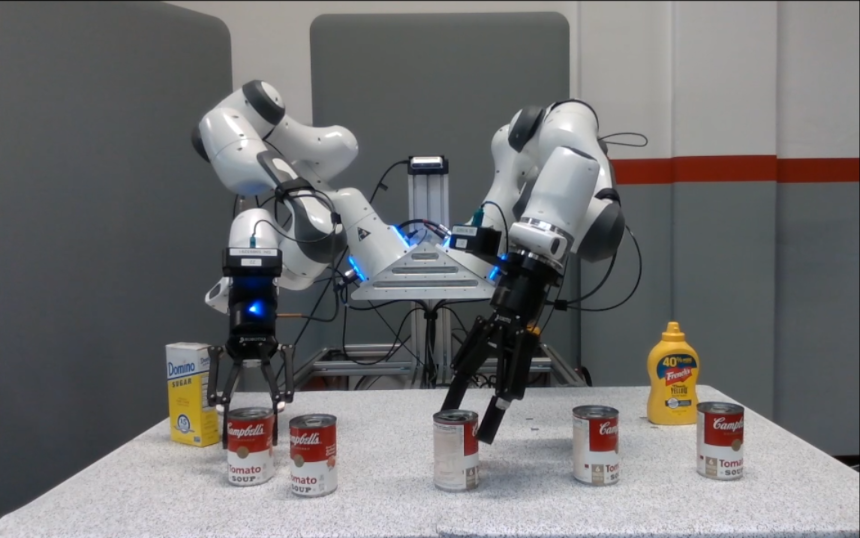}};
    \node[anchor=west, right=of pt4.east]  (pt5) {\includegraphics[width=.19\textwidth]{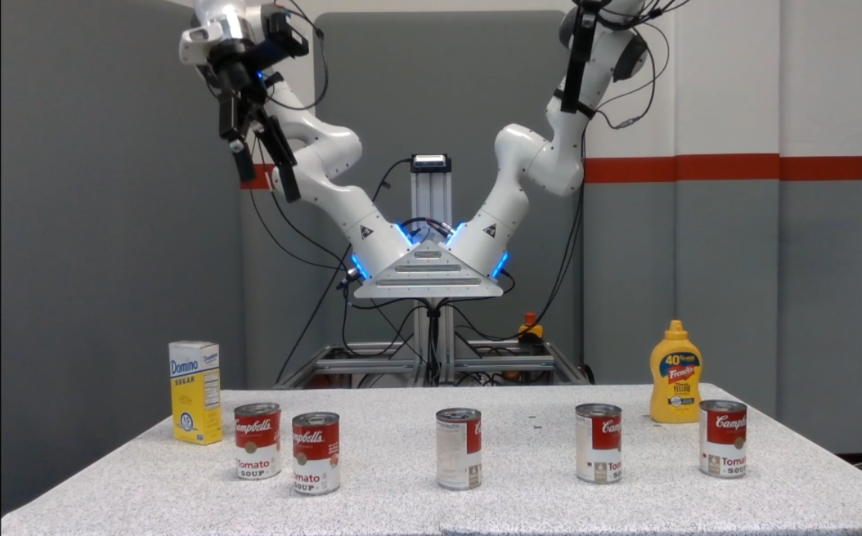}};

    \node[anchor=north, below=of pt1.south] (pb1) {\includegraphics[width=.19\textwidth]{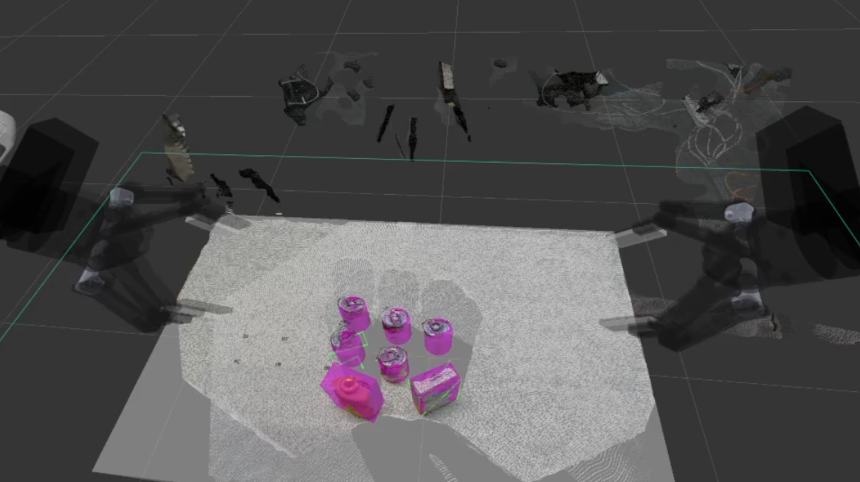}};
    \node[anchor=west, right=of pb1.east]   (pb2) {\includegraphics[width=.19\textwidth]{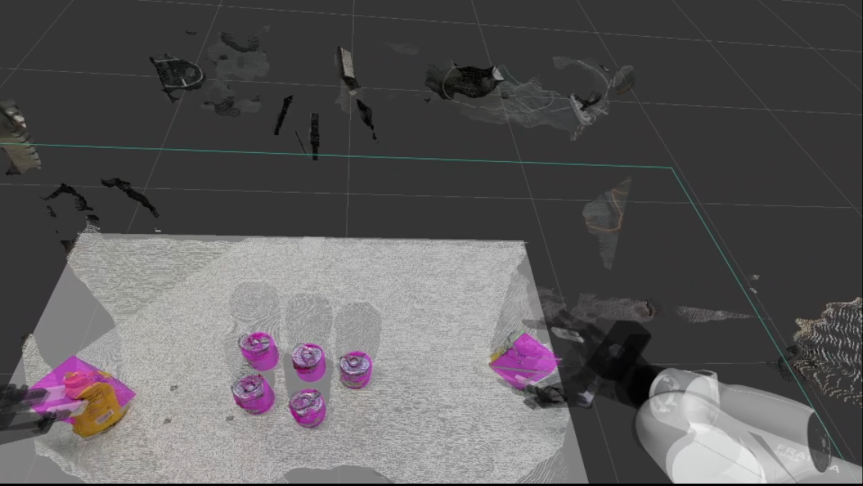}};
    \node[anchor=west, right=of pb2.east]   (pb3) {\includegraphics[width=.19\textwidth]{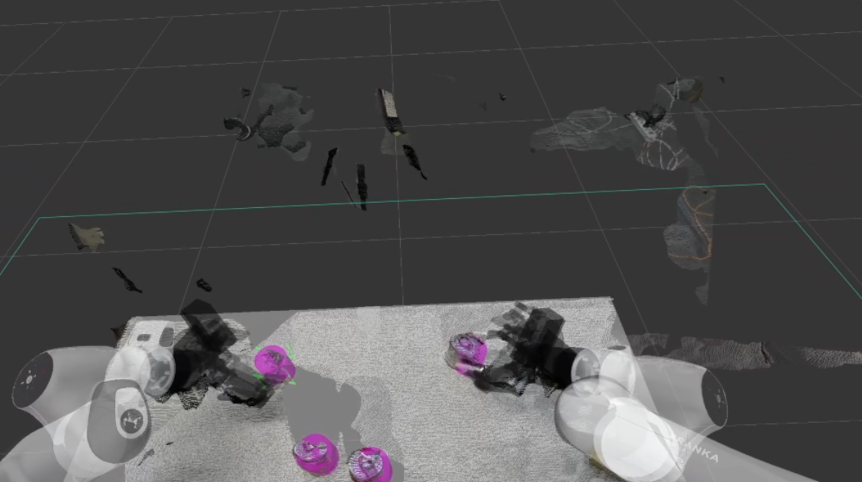}};
    \node[anchor=west, right=of pb3.east]   (pb4) {\includegraphics[width=.19\textwidth]{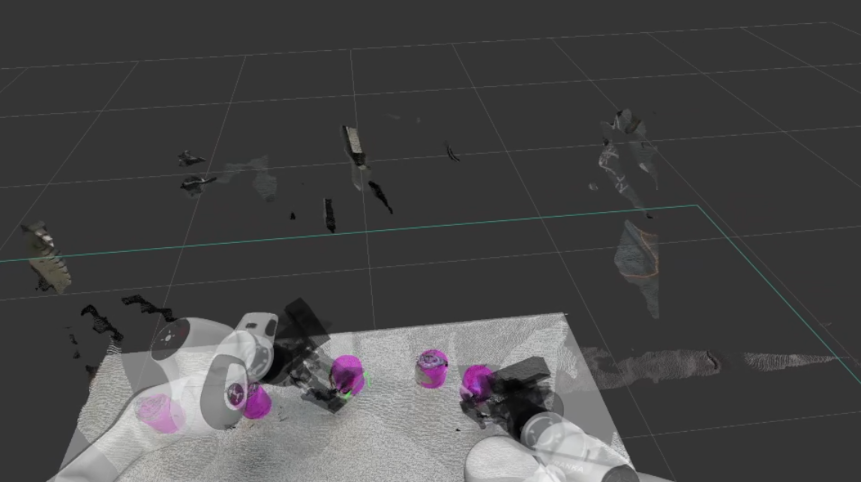}};
    \node[anchor=west, right=of pb4.east]   (pb5) {\includegraphics[width=.19\textwidth]{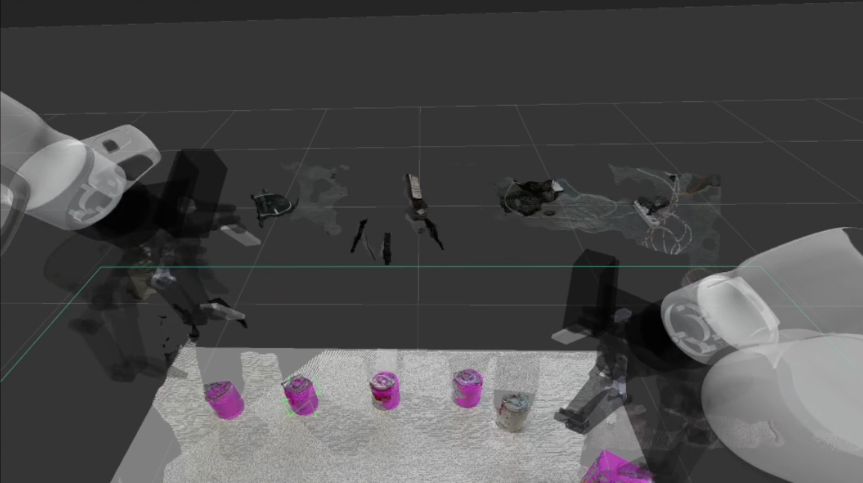}};    
  \end{tikzpicture}
  \caption{Pictures from the experiment with the real robot. The top pictures show the robot while the bottom pictures show the operator view in \acronym{VR}. The transparent end-effectors in the operator view correspond to the target pose set by the operator.
  The leftmost pictures show the initial object configuration and the rightmost pictures the final result after cleaning the table.}
  \label{fig:robot_experiment}
\end{figure*}

\textbf{Constraint Cost Prediction}:
\hyperref[tab:constraints]{Table~\ref*{tab:constraints}} shows an overview of the models used for predicting the constraint costs. All models are fully connected Neural Networks with \textit{LeakyReLU} nonlinear layers. Since false negative predictions~(e.g.,~falsely predicting a configuration is collision-free) are much more severe than false positive predictions, we use a quantile loss for training were we punish false negative predictions stronger by a factor of 2. We split the collision prediction into multiple modules since static collisions can be predicted solely based on the robot's configuration while dynamic collisions require knowledge about the poses of the objects. Dynamic collisions are predicted for each object individually before aggregating the prediction by taking the maximum as the final output~(e.g., when object is in a collision the overall predicted collision costs will be positive). The pre-grasp constraint has fewer features compared to the other predictors because it only requires knowledge about the pose of the end-effector in the object frame as well as the object's parameters.
As metric for the pre-grasp constraint we defined a coordinate frame between the two finger pads of the gripper. We consider an end-effector pose to be a valid pre-grasp pose if this coordinate frame overlaps with the object. Please note that this also includes configurations where the end-effector collides with the object. However, such configurations are not problematic since they violate the dynamic collision constraint and are therefore not feasible under consideration of all activate constraints. All constraint cost predictors were trained with the binary cross entropy loss function. The reason is we are not interested in approximating the exact constraint costs but rather treat the approximation as binary classification problem where the system only has to get the decision on if the constraint is satisfied or not right.

\subsection{Real-Time Teleoperation}
The aim of the experiment with the real robot was to evaluate whether our system can provide adequate assistance for real-time teleoperation. The task was to clean up a table by picking up $7$ objects at the center of the table and moving them to the table boundaries, as depicted in \hyperref[fig:robot_experiment]{Figure~\ref*{fig:robot_experiment}}.
All objects were from the \acronym{YCB}-video dataset \cite{xiang2018posecnn}. Since we trained our constraint cost predictors on box- and cylinder-shaped objects we selected objects from the dataset which roughly had such a shape. For the evaluation, we counted the number of successfully picked up objects and the overall task completion time and compared it against a baseline where the system did not provide any assistance. Hence, the operator directly controlled the 6D poses of the end-effectors with the \acronym{VR} controllers. In total $6$ participants performed the experiment (baseline and our method). All participants had some experience with teleoperating the robot but no experience with the assistance system. Before starting the experiments, the operators were allowed to do some test trials to get used to the assisted method as well as the baseline. Snapshots from the robot experiments are depicted in \hyperref[fig:robot_experiment]{Figure~\ref*{fig:robot_experiment}}. For a more in-depth look, we encourage interested readers to view the accompanying video.

\begin{table}[h!]
  \centering
  \begin{tabular}{| c | c | c|} 
   \hline   
   Metric & Our Method & Baseline \\ 
   \hline
   Successful trials & $\boldsymbol{12/12}$ & $10/12$ \\ 
   Avg. task completion time & $90.5$s & $\boldsymbol{87.6}$s \\ 
   Grasp attempts & $\boldsymbol{84/90}$ & $84/92$ \\ 
   Repositioned objects & $\boldsymbol{84/84}$ & $77/84$ \\
   \hline
  \end{tabular}
  \caption{Results on the table cleaning task. Our method allows the operator to reliably grasp and reposition the objects while being only slightly slower than the baseline.}
  \label{tab:experiments}
\end{table}

The robot command generated by our system was based on $1024$ sampled configurations. The commands were filtered with a window size of 3. Without filtering, the robot was more reactive but the resulting motions were a bit jerky. We consider it future work to find the best balance between having smooth robot motions and having a highly reactive robot system. Overall, we performed $12$ trials with each method, respectively. The results are depicted in \hyperref[tab:experiments]{Table~\ref*{tab:experiments}}. All participants managed to perform the task with the assisted method, but sometimes objects slipped because they were not grasped properly. We counted $6$ failed grasp attempts for the assisted method, for $84$ successful grasps. For the baseline, the participants only succeeded in $10$ out of $12$ trials. Failure cases were an emergency stop because of a collision and a case where the operator maneuvered the robot into joint limits from where the robot could not recover. The failed grasps can mainly be attributed to the grasp not being centered on the object, causing it to slip when attempting to lift it. Overall, the average task completion time for the assisted method was a bit higher due to the filtering of the output commands which tends to slow down the robot. 

\subsection{Discussion}
The feedback we received from the participants was two-fold. While the assisted method generally performed better in terms of reliably grasping objects, operators needed to be more patient as the robot occasionally paused before an object until the system found a feasible grasp pose. This brief waiting period particularly bothered the more experienced operators. In contrast, less experienced operators appreciated the method’s ability to automatically handle potential collisions and found it helpful that they only needed to provide a rough indication of how they wanted an object to be grasped, without controlling every aspect of the motion.

Based on the feedback, we conclude that the assisted method can improve the teleoperation experience, though there is still significant room for improvements. Currently, we sample potential target configurations in joint space which leads to many unrealistic configurations. Switching to sampling in task-space might help here, but can also slow down the method since it requires computing the pseudo-inverse of the Jacobian matrix before sampling. Additionally, it is crucial that the assistance method is transparent to the operator, allowing him/her to understand the system's current target and to assess whether the system can provide assistance in the current situation. Lastly, enabling operators to seamlessly switch between assistance and pure teleoperation mode is important to maintain the sense of agency.

Lastly, the limited number of participants prevents us from drawing definitive conclusions about the effectiveness of our method. While the feedback from participants suggests that our method could be useful, further experiments are necessary to draw more concrete conclusions.

\section{Conclusion and Future Work}
In this paper, we presented a concept for assisted teleoperation/shared autonomy that uses Neural Networks for approximating task-relevant constraints. We interpreted the assisted teleoperation problem as an optimization problem. Instead of solving the problem in every iteration, we presented a concept for locally approximating it via sampling and constraint cost prediction. The sampling-based selection of a robot configuration allows us to minimize the delay between receiving the operator input and sending a control command to the robot, which allows for a smooth operation of the robot.

In future work, we want to add more constraints in order to support more sophisticated manipulation tasks such as insertion. The constraint predictors can also be improved. For instance, currently the feasible pre-grasp constraint cost predictor is only trained on simple objects (cylinders and boxes). Furthermore, the code of the framework could be optimized to achieve even higher frame rates.



\bibliography{references}{}
\bibliographystyle{ieeetr}




\end{document}